\documentclass[preprint,review,10pt,authoryear]{elsarticle}
\pdfoutput=1 
\usepackage{lineno,hyperref}
\modulolinenumbers[5]
\usepackage{algorithm,algpseudocode}
\usepackage{caption}
\usepackage{url}
\usepackage{footnote} 
\usepackage{epstopdf}
\usepackage{epsfig}
\usepackage{graphicx}
\usepackage{array}
\usepackage{amsfonts,epsfig,array,graphicx,amssymb,epstopdf}
\usepackage[]{natbib}
\usepackage{cite}

\usepackage{times}
\usepackage{graphicx}
\usepackage{amsmath}
\usepackage{amssymb}
\usepackage{booktabs}
\usepackage{multirow}
\usepackage{epstopdf}
\usepackage{caption}
\usepackage{subcaption}
\usepackage{tabularx}
\usepackage{algorithm,algpseudocode}
\usepackage{lineno,hyperref}
\usepackage{float}
\usepackage{pdflscape}
\usepackage{mathrsfs}
\usepackage{longtable}
\usepackage{ltablex}
\usepackage{lineno,hyperref}
\usepackage{csquotes}

\usepackage{algpseudocode}

\algdef{SE}[DOWHILE]{Do}{doWhile}{\algorithmicdo}[1]{\algorithmicwhile\ #1}%

\journal{Journal of Expert Systems with Applications}

\begin{document}

\begin{frontmatter}

\title{Fingertip Detection and Tracking for Recognition of Air-Writing in Videos}


\author{$^*$Sohom Mukherjee$^a$, Sk. Arif Ahmed$^b$, Debi Prosad Dogra$^c$, Samarjit Kar$^d$ and Partha Pratim Roy$^e$}

\address{Department of Electrical Engineering,\\National Institute of Technology Durgapur, Durgapur-713209, India$^{a}$\\ Department of Mathematics,\\National Institute of Technology Durgapur, Durgapur-713209, India$^{b,d}$\\ School of Electrical Science,\\Indian Institute of Technology Bhubaneswar, Bhubaneswar-751013, India$^c$\\ Department of Computer Science and Engineering,\\Indian Institute of Technology Roorkee, Roorkee-247667, India$^e$\\

Email addresses:\ sohom31011997@gmail.com$^a$, arif.1984.in@ieee.org$^b$, dpdogra@iitbbs.ac.in$^c$, samarjit.kar@maths.nitdgp.ac.in$^d$, proy.fcs@iitr.ac.in$^e$}

\cortext[mycorrespondingauthor]{Corresponding author. Tel.: +91
9434184926\\\textit{Email address:} \texttt{sohom31011997@gmail.com} (*Sohom Mukherjee)} 



\begin{abstract}
Air-writing is the process of writing characters or words in free space using finger or hand movements without the aid of any hand-held device. In this work, we address the problem of mid-air finger writing using web-cam video as input. In spite of recent advances in object detection and tracking, accurate and robust detection and tracking of the fingertip remains a challenging task, primarily due to small dimension of the fingertip. Moreover, the initialization and termination of mid-air finger writing is also challenging due to the absence of any standard delimiting criterion. To solve these problems, we propose a new writing hand pose detection algorithm for initialization of air-writing using the Faster R-CNN framework for accurate hand detection followed by hand segmentation and finally counting the number of raised fingers based on geometrical properties of the hand. Further, we propose a robust fingertip detection and tracking approach using a new signature function called distance-weighted curvature entropy. Finally, a fingertip velocity-based termination criterion is used as a delimiter to mark the completion of the air-writing gesture. Experiments show the superiority of the proposed fingertip detection and tracking algorithm over state-of-the-art approaches giving a mean precision of 73.1 \% while achieving real-time performance at 18.5 fps, a condition which is of vital importance to air-writing. Character recognition experiments give a mean accuracy of 96.11 \% using the proposed air-writing system, a result which is comparable to that of existing handwritten character recognition systems.
\end{abstract} 

\begin{keyword}
\texttt{Air-writing, Hand pose detection, Fingertip detection and tracking, Handwritten character recognition, Human-computer interaction (HCI)}
\end{keyword}

\end{frontmatter}

\section{Introduction}

With the emergence of virtual and augmented reality, the need for the development of natural human-computer interaction (HCI) systems to replace the traditional HCI approaches is increasing rapidly. In particular, interfaces incorporating hand gesture-based interaction have gained popularity in many fields of application viz. automotive interfaces \citep*{ohn2014hand}, human activity recognition \citep*{rohrbach2016recognizing} and several state-of-the-art hand gesture recognition approaches have been developed \citep*{molchanov2015hand,rautaray2015vision}. However, hand motion gestures as such are not sufficient to input text. This necessitates the need for the development of touch-less air-writing systems which may replace touch and electromechanical input panels leading to a more natural human-computer interaction (HCI) approach.

A vision-based system for the recognition of mid-air finger-writing trajectories is not a new problem and substantial work has been done in the past two decades. One of the early works by Oka et al. \citep*{oka2002real} used a sophisticated device with an infrared and color sensor for fingertip tracking and recognition of simple geometric shapes trajectories. In \citep*{amma2012airwriting}, inertial sensors attached to a glove were used for continuous spotting and recognition of air-writing. Recently, Misra et al. \citep*{misra2017vision} have developed a hand gesture recognition framework that can recognize letters, numbers, arithmetic operators as well as 18 printable ASCII characters using a red marker placed on the tip of the index finger for fingertip detection. In spite of satisfactory performance in terms of accuracy of character trajectory recognition, the above approaches using cumbersome motion sensing and tracking hardware devices impose many behavioral constraints on the user. For example, wearing data gloves on the hand may change the user's natural handwriting pattern and is often considered as an undesirable burden by many users. During the past few years, several research works have been carried out addressing the problem of air-writing recognition using depth sensors such as Microsoft Kinect and special hardware namely the Leap Motion Controller from Leap Motion\footnote{\url{https://www.leapmotion.com/}}, for input. Chang et al. \citep*{chang2016spatio} have proposed a Spatio-Temporal Hough Forest for the detection, localization and recognition of mid-air finger-writing in egocentric depth video.  Chen et al. \citep*{chen2016air} have developed a robust system for air-writing recognition using the Leap Motion Controller for marker-free and glove-free finger tracking. However, the high cost and limited availability of such sophisticated hardware to the majority of users render them unsuitable for real-world applications.

The work that is closest to ours is by Huang et al. \citep*{huang2016pointing} where they use a two stage CNN-based fingertip detection framework for recognition of air-writing in egocentric RGB video. However, capturing egocentric video requires head-mounted smart cameras or mixed reality headsets such as Google Glass, Microsoft HoloLens and Facebook Oculus, which may not be available to all users. Therefore, to make our system more general, we use video inputs from a standard laptop camera or a web-cam for the air-writing application. This makes the task even more challenging due to the presence of the face in the video frames, which being a moving object of similar skin tone as the hand, makes the detection of the hand and hence the fingertip much more complicated.

To address these issues we propose a new system for air-writing recognition in videos using a standard laptop camera or a web-cam for the video input. The key challenges in this task are: (1) writing hand pose detection for initialization of air-writing; (2) accurate fingertip detection and tracking in real-time; (3) recognition of air-writing character trajectory. The task is fairly difficult due to several factors such as hand shape deformation, fingertip motion blur, cluttered background and variation in illumination.

To address the aforementioned challenges, our work makes the following contributions:

\begin{itemize}
 \item We propose a new writing hand pose detection algorithm for the initialization of air-writing using the Faster R-CNN framework for accurate hand detection followed by hand segmentation and finally counting the number of raised fingers based on geometrical properties of the hand.

 \item We propose a robust fingertip detection approach using a new signature function called distance-weighted curvature entropy.

 \item We propose a fingertip velocity-based termination criterion which serves as a delimiter and marks the completion of the air-writing gesture.
\end{itemize}

The rest of the paper is organized as follows. Section \ref{related work} provides a review of the related work. Section \ref{proposed algorithm} presents the proposed air-writing recognition algorithm detailing its four stages namely writing hand pose detection, fingertip detection, fingertip tracking and trajectory generation and character recognition. In Section \ref{experiments}, details of the implementation, dataset and experimental analysis are presented. Finally, in Section \ref{conclusion}, conclusions are drawn and the scope for future development of the proposed approach is discussed.


\section{Related Work} \label{related work}
Recognizing mid-air finger-writing is a difficult and open problem in computer vision, due to the various challenges outlined in the previous section. We review the previous research works related to this work as follows: (1) fingertip detection and tracking; (2) air-writing recognition; and (3) the existing hand datasets related to air-writing recognition.

\subsection{Fingertip Detection and Tracking}

Fingertips detection and tracking has been an active topic in the fields of HCI and augmented reality (AR) using color as well as depth cameras. A model-based approach for 3D hand tracking \citep*{de2011model, tang2017latent} may be used as a preceding step for fingertip detection, but these approaches involve high computational cost and require a large amount of training data. This makes them unsuitable for our real-time application. In the model-less approach, the hand silhouette is first segmented using color, depth or motion cues and then the fingertips are detected from the extracted binary hand mask. Liang et al. \citep*{liang20123d} used a distance metric from hand palm to the contour furthest points to localize candidate fingertip points. Krejov and Bowden \citep*{krejov2013multi} extended the distance concept employing it with the natural structure of hand using a geodesic distance. This improved the localization of fingertips in hand configurations where previous methods failed. In \citep*{lee2007handy}, the authors used the contour curvature as a cue to detect fingertips, exploiting the fact that fingertips are high curvature points, compared to other parts of the hand. 

However, all the above approaches suffer from the drawback that the fingertip detection depends largely on the hand segmentation, and suffer severely from poor segmentation results while making use of solely color, depth and motion cues \citep*{zhang2013new}. The problem can be solved by accurate detection of hands as a preceding step for hand segmentation. In \citep*{mittal2011hand}, a skin-based detector, a deformable part models (DPM)-based hand shape detector and a context detector are combined to detect hands in still images, but this method is time consuming for real-time applications due to the sliding window approach. In \citep*{li2013pixel}, hand detection and pixel-level segmentation from egocentric RGB videos using local appearance features have produced good results, but the performance is affected by variation in illumination.

Recent advances in deep learning based methods are giving excellent results on general object detection tasks. However hand detection remains a challenging task and numerous works have been done in recent times. The Region-based CNN (R-CNN) framework is applied in \citep*{bambach2015lending} to detect hands in egocentric videos. This method achieves promising performance in complex environments but is slow due to CNN-based feature computation for redundant overlapping region proposals. A recent work by Deng et al. \citep*{deng2018joint} proposes a CNN-based joint hand detection and rotation estimation framework, using on an online de-rotation layer embedded in the network. In \citep*{roy2017deep}, a two-step framework is proposed for detection and segmentation of hands using a Faster R-CNN based hand detector followed by a CNN based skin detection technique. An experimental survey of existing hand segmentation datasets, state-of-the art methods as well as three new datasets for hand segmentation can be found in \citep*{khan2018analysis}. Deep learning based methods have been extended to fingertip detection in \citep*{huang2016pointing}. They use a two-stage framework consisting of Faster R-CNN based hand detection followed by CNN-based fingertip detection. More recently, Wu et al. \citep*{wu2017yolse} proposed a framework called YOLSE (You Only Look what You Should See), that uses a heatmap-based fully convolution network for multiple fingertip detection from single RGB images in egocentric view.

Tracking a small object such as the fingertip from an image accurately and robustly remains a great challenge and to the best of our knowledge no standard method for fingertip tracking has yet been proposed. Mayol et. al. \citep*{mayol2004interaction} and Kurata et. al. \citep*{kurata2001hand} have used template matching and mean-shift respectively for hand tracking in constrained environments, from images captured using a wearable camera. \citep*{stenger2006model} gives a tracking framework, which is applied to the recovery of three-dimensional hand motion from an image sequence using a hierarchical Bayesian filter. In \citep*{kolsch2004fast}, a fast tracking method has been proposed for non-rigid and highly articulated objects such as hands. It uses KLT features along with a learned foreground color distribution for tracking in 2D monocular videos. However, these methods cannot deal with long-term continuous tracking problems such as change in object appearance and object moving in and out of the frame, since they are designed and evaluated for short video sequences. The Tracking-Learning-Detection (TLD) framework \citep*{kalal2012tracking} has been proposed for long-term tracking. TLD works well when there are frequent variations in the appearance of an object, but is slow compared to many other tracking algorithms. Therefore, long-term tracking of small objects remains a challenging task.

\subsection{Air-Writing Recognition}
Some methods have been proposed in the literature for the recognition of air-writing, treating it as a spatial-temporal signal. In \citep*{chen2016air}, various modifications of Hidden Markov Models (HMMs) have been used for the recognition of air-writing generated using Leap Motion Controller. An attention-based model, called attention recurrent translator has been used in \citep*{gan2017air} for in-air handwritten English word recognition, which gives performance comparable to the connectionist temporal classification (CTC). Kane and Khanna \citep*{kane2017vision} have proposed an equipolar signature (EPS) technique for a vision-based mid-air unistroke character input framework that is resistant to variations in scale, translation and rotation. Some recent works have also addressed the problem of 3D air signature recognition and verification \citep*{behera2017analysis, behera2018fast}. \citep*{behera2018fast} presents a method for analyzing 3D air signatures captured using the Leap Motion Controller, with the help of a new geometrical feature extracted from the convex hull enclosing a signature. 
 
In recent years, deep convolutional neural networks (CNNs) have achieved great success in handwritten character recognition, beating benchmark performances by wide margins \citep*{ciregan2012multi, cirecsan2015multi}. The multi-column deep neural network proposed in \citep*{ciregan2012multi} reaches near-human performance on the MNIST handwritten digits dataset. A variation of CNNs called DeepCNet, first proposed by Graham \citep*{graham2013sparse}, won first place in the ICDAR 2013 Chinese Handwriting Recognition Competition \citep*{yin2013icdar}. Further advances in CNN architectures namely DropWeight \citep*{xiao2017design} and new training strategies such as DropDistortion \citep*{lai2017toward} and DropSample \citep*{yang2016dropsample} have led to improved performance of CNNs in online handwritten character recognition tasks.

\subsection{Related Datasets}
At present, there isn't any benchmark dataset in the area of finger air-writing research, to evaluate the performance of the methods. Recently, some datasets have been released for the recognition of air-writing in egocentric view. The EgoFinger dataset \citep*{huang2016pointing} containing 93,729 labelled RGB frames from 24 egocentric videos and captured from 24 different individuals, is the most extensive among such datasets. Other datasets for the recognition of egocentric hand gesture include the EgoHands dataset \citep*{bambach2015lending}, containing 48 videos of egocentric interactions between two people, with pixel-level ground-truth annotations for 4,800 frames and 15,053 hands. RGB-D datasets are comparatively more numerous. The EgoGesture \citep*{ zhang2018egogesture} is a multi-modal large scale dataset for egocentric hand gesture recognition containing 2,081 RGB-D videos, 24,161 gesture samples and 2,953,224 frames acquired from 50 distinct subjects. In \citep*{suau2014real} a real-time fingertip detection method has been proposed with fingertip labeled RGB-D dataset. However, none of these datasets are suitable for our proposed air-writing application using web-cam video input. 

\begin{figure*}[t]
\centering
\includegraphics[scale=0.5]{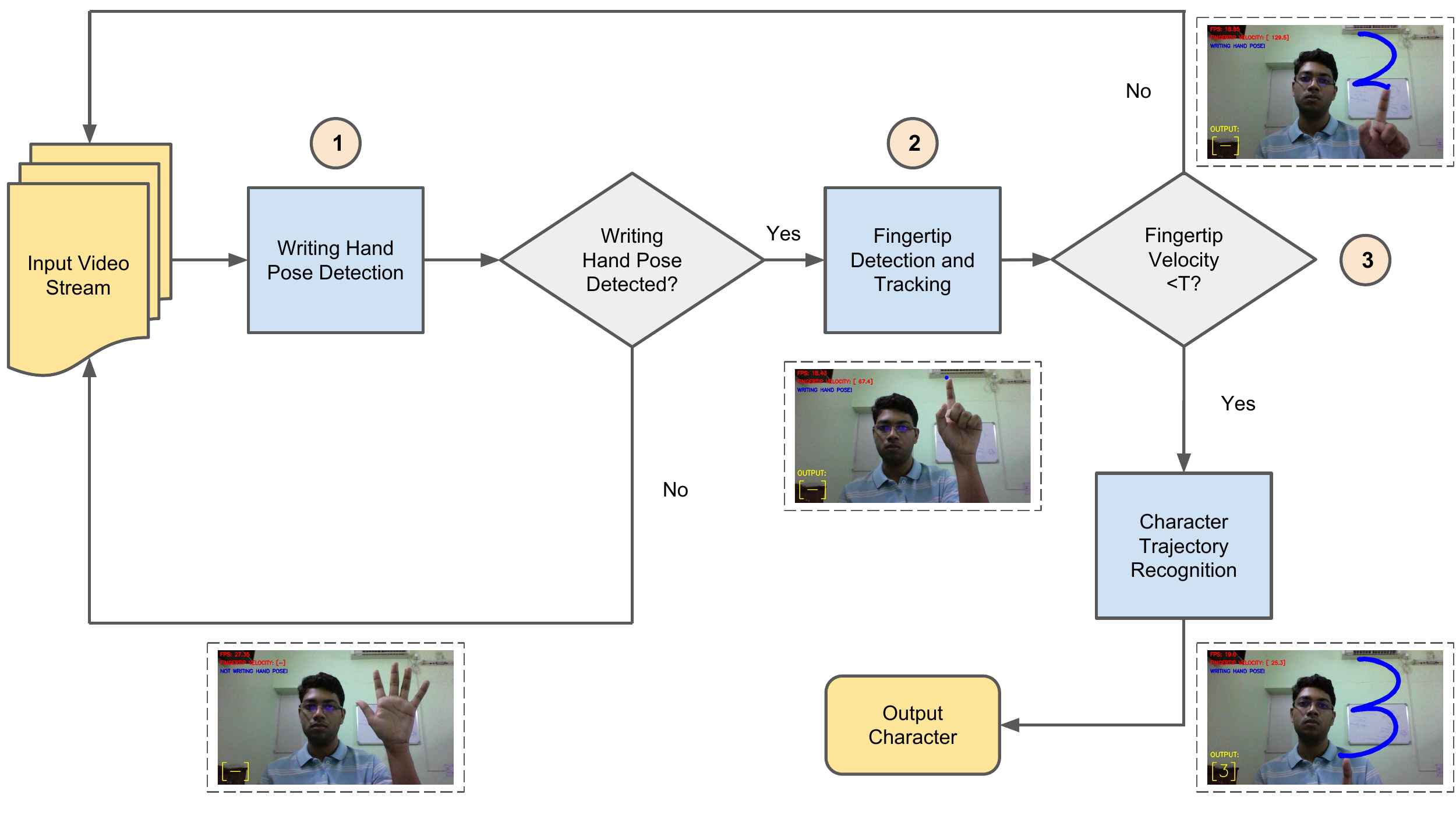}
\caption{Flowchart for the proposed air-writing recognition system. The three main contributions in the work have been highlighted by numbering them in the order in which they appear is the pipeline.}
\label{fig:fc_0}
\end{figure*}

\section{Proposed Algorithm} \label{proposed algorithm}

A flowchart for the proposed air-writing recognition system is shown in Figure \ref{fig:fc_0}. The proposed algorithm can be broken into the following four components: (1) writing hand pose detection; (2) fingertip detection in each frame; (3) fingertip tracking and character trajectory generation; (4) recognition of air-writing characters.

\subsection{Writing Hand Pose Detection}

The detection of writing hand pose and its recognition from other gestures is an integral step towards air-writing initialization since, unlike conventional handwriting which has the pen-down and pen-up motion, air-writing does not have such a delimited sequence of writing events. In this work, we define the writing hand pose to be a single raised finger with the assumptions that the finger is free from occlusions and is not manipulating any object. We detect a writing hand pose and discriminate it from a non-writing hand pose by counting the number of raised fingers. To this end we propose a four-fold approach consisting of hand detection, hand region segmentation, localization of the hand centroid, followed by utilizing geometrical properties of the hand to count the number of raised fingers.

\subsubsection{Hand Detection} \label{hand detection}

The Faster R-CNN (FRCNN) \citep*{ren2015faster} framework has been used for hand detection. Faster R-CNN is a state-of-the-art object detection algorithm that removes dependency on external hypothesis generation methods, as in case of Fast R-CNN \citep*{7410526}. The first step involves using the output of an intermediate layer of a CNN pretrained for the task of classification (called base network) to generate a convolutional feature map. The original Faster R-CNN used VGG-16 network \citep*{simonyan2014very} pretrained on ImageNet, but since then many different networks with a varying number of weights have been used. The base network returns a convolutional feature map that forms the input to the Region Proposal Network (RPN). The RPN is a fully convolutional network that takes the reference bounding boxes (anchors) and outputs a set of object proposals. This is followed by region of interest (RoI) pooling for extraction of features from each of the object proposals. Finally, we use these features for classification using a region-based convolutional neural network (R-CNN). After putting the complete FRCNN model together, we train it end-to-end as a single network, with four losses, RPN classification loss, RPN regression loss, R-CNN classification loss and R-CNN regression loss. The details of implementation and training of the proposed Faster R-CNN based hand detector is given in Section \ref{hand experiment}.

\begin{figure*}[t]
\centering
\includegraphics[scale=0.5]{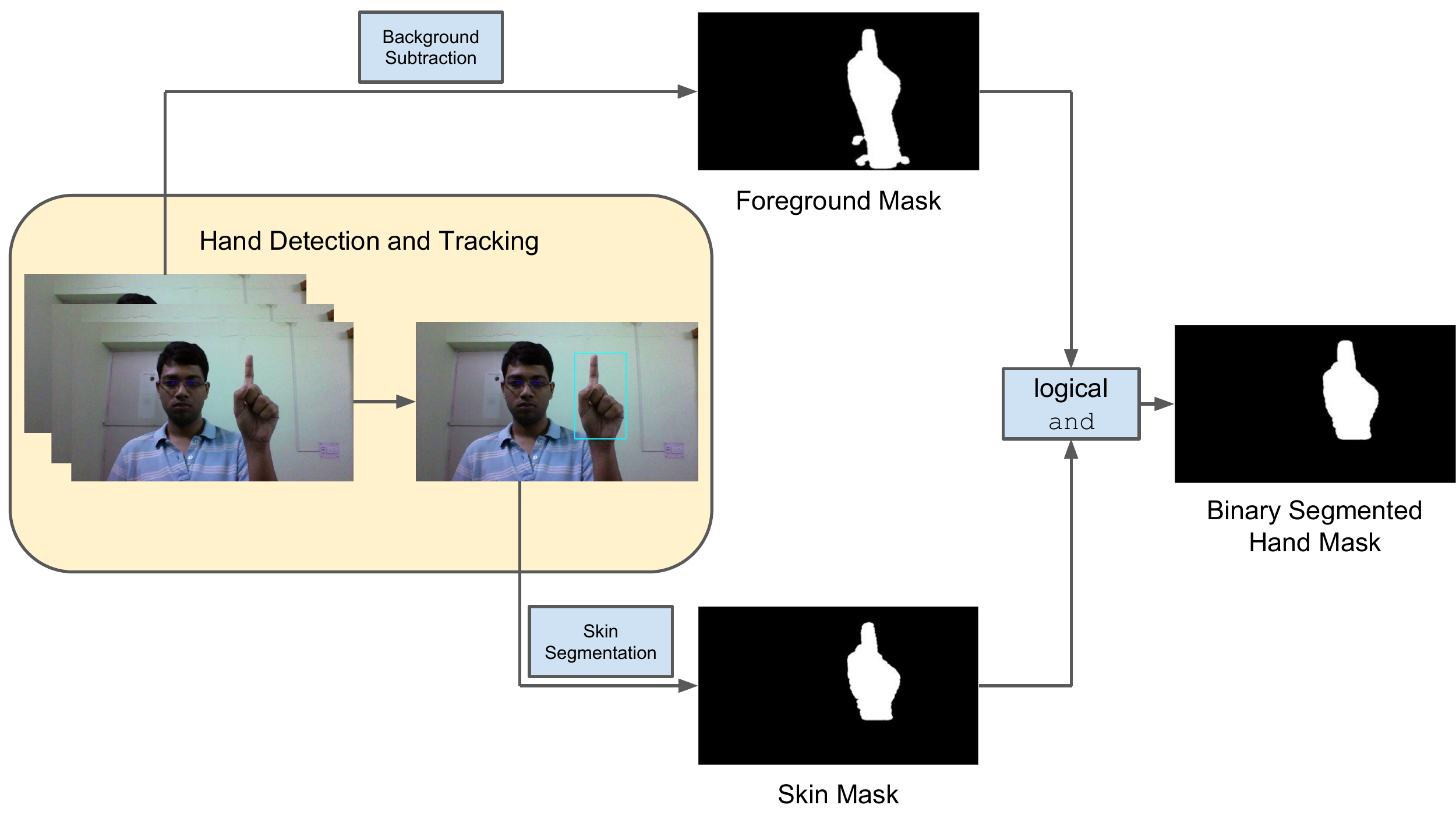}
\caption{Hand detection and segmentation.}
\label{fig:hand-detection-segmentation}
\end{figure*}

\subsubsection{Hand Region Segmentation}
Once we have accurately detected the hand using the above technique, hand region segmentation is achieved using a two-step approach viz. skin segmentation and background subtraction and the final binary hand image is obtained as an aggregation of the two. In spite of the success of semi-supervised segmentation algorithms namely GrabCut for hand segmentation \citep*{bambach2015lending}, we find that such algorithms are computationally expensive and hence not suitable for our purpose. The proposed algorithm is found to work well in real-time and gives fairly accurate segmentation results. Figure \ref{fig:hand-detection-segmentation} shows a schematic diagram for the proposed hand segmentation algorithm. 

\textit{Skin Segmentation:} We propose a skin color filtering approach based on a static skin color model in the YCbCr color space for skin segmentation. Although skin colors vary considerably across races, it has been observed that the skin chrominance has only a small range across different skin types, while the skin luminance varies considerably. This forms the basis of the static skin color model \citep*{phung2005skin}. Based on  extensive experimentation \citep*{phung2005skin}, we extracted the narrow band $77\leq Cb \leq 127 \operatorname{\,and\,} 133\leq Cr \leq 173$ as our static skin color model and discarded the Y component to exclude the luminance information. A frame of the video sequence acquired by the web-cam is in RGB color space, and we can convert it to YCbCr color space using the following equation \citep*{wu2016robust}:

\begin{equation}
\begin{bmatrix}Y \\ Cb \\ Cr \\ 1 \end{bmatrix} =
\begin{bmatrix}
0.2990 & 0.5870 & 0.1140 & 0 \\
-0.1687 & -0.3313 & 0.5000 & 128 \\
0.5000 & -0.4187 & -0.0813 & 128 \\
0 & 0 & 0 & 1 \\
\end{bmatrix}
\cdot
\begin{bmatrix}R \\G \\B \\ 1 \end{bmatrix}
\end{equation}

We construct a skin filter based on the YCbCr static skin color model and filter the candidate hand bounding box, obtained from detection phase to get the binary skin mask $I_{h1}$ as follows:

\begin{equation}
I_{h1}(x,y)=\begin{cases}1 & \operatorname{\,if\,} 77\leq Cb \leq 127 \operatorname{\,and\,} 133\leq Cr \leq 173
\\ 0 & \operatorname{\,otherwise\,}\end{cases}
\end{equation}

Skin color filtering method is chosen for skin segmentation over other methods namely Gaussian classifiers and the Bayesian classifier, in spite of their slightly better performance, since they increase the computational load significantly and hence are not suitable for real-time applications.

\textit{Background Subtraction:} Since accurate hand detection using the Faster R-CNN based hand detector followed by skin color filtering in the candidate hand bounding box gives a reasonably good segmentation result, the background subtraction step is used only to remove any skin colored object (not a part of the hand) that may be present inside the detected hand bounding box. We use the adaptive Gaussian Mixture Model (GMM) based background subtraction algorithm proposed in \citep*{zivkovic2004improved} for this purpose. The main advantage of GMM is that it can reach real-time processing. Let the binary foreground mask obtained from this step be $I_{h2}$.

The final binary hand image is obtained by a $\operatorname{\,logical \, and\,}$ operation between the binary skin mask $I_{h1}$ and the binary foreground mask $I_{h2}$ as follows:

\begin{equation}
I_{h} = I_{h1} \wedge I_{h2}
\end{equation}

The above approach using an aggregation of two segmentation algorithms gives us an accurate hand segmentation result which is free from noise due to moving objects in the scene which are not the hand as well as any other object inside the detected hand bounding box which is skin colored but does not belong to the hand. The segmented hand image $I_{h}$ is then filtered applying the morphological operations of dilation $\oplus$ and erosion $\ominus$, as described in equation (\ref{eq:4}), where $I_{h}$  is the segmented hand image, $I_{hf}$ is the resulting filtered image and $K$ is a circular disk structuring element of radius $3$ with the anchor at its center.

\begin{eqnarray} \label{eq:4}
I_{hf} = ((I_{h}\oplus K) \ominus K) \oplus K
\end{eqnarray}

These morphological operations are employed in order to fill spaces in the foreground object (hand) with dilation, and then clean any background noise with an erosion. Finally, the hand silhouette is regularized using dilation. Lastly, we sort all connected regions in $I_{hf}$ by area, and extract the connected region having the maximum area as the hand region.

\begin{figure}[t]
 
\begin{subfigure}{0.5\textwidth}
\includegraphics[scale=1, width=1\linewidth]{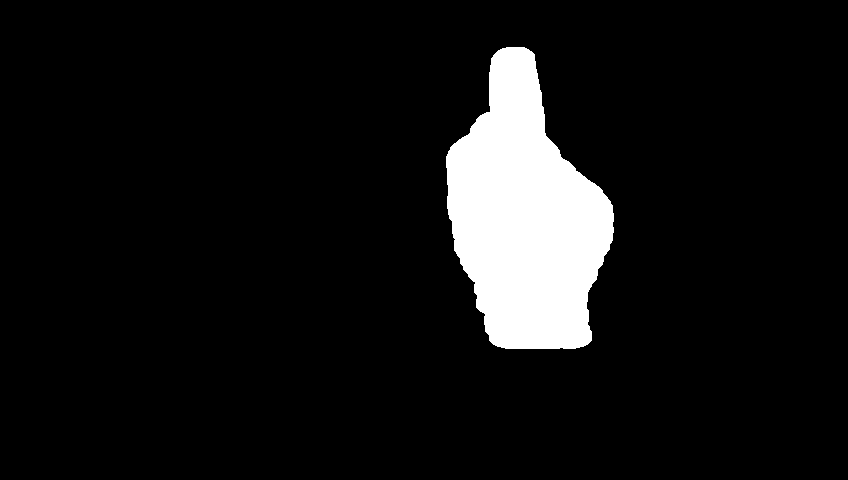} 
\caption{}
\label{fig:handmask}
\end{subfigure} 
\begin{subfigure}{0.5\textwidth}
\includegraphics[scale=1, width=1\linewidth]{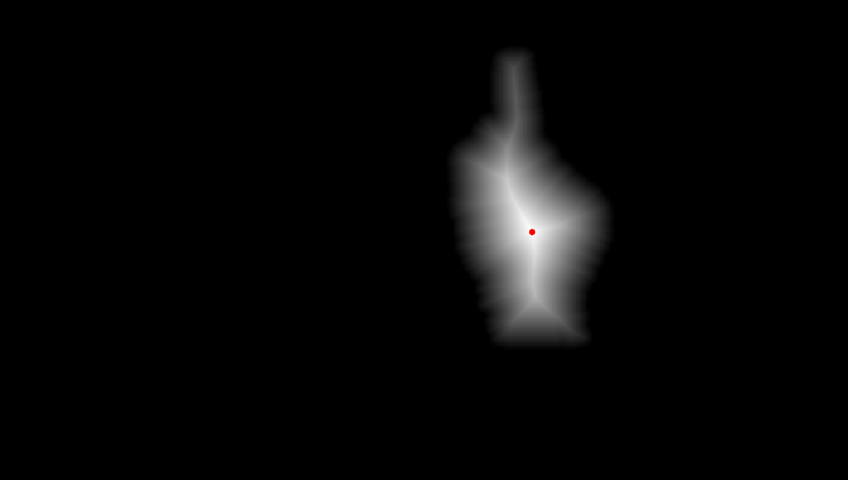}
\caption{}
\label{fig:dthand}
\end{subfigure}

\caption{Hand centroid localization using distance transform. (a) Binary hand image. (b) Distance transform (the pixel having the maximum intensity is highlighted with a red marker).}
\label{fig:distancetransform}
\end{figure}

\subsubsection{Hand Centroid Localization}

Since the accurate determination of the centroid of the hand plays a crucial role in the following steps, we employ two algorithms to find the initial estimates of the centroid and the final centroid is calculated as the mean of the two.

The method of distance transform is used to get the first estimate of the centroid $(x_{c1}, y_{c1})$. In the distance transform image, each pixel is represented by its distance from the nearest boundary pixel. Figure \ref{fig:distancetransform} shows an example of distance transform. Figure \ref{fig:handmask} shows a binary hand image and Figure \ref{fig:dthand} shows the distance transform image of the former. The Euclidean distance has been used to measure the distance between a pixel and its nearest boundary pixel.  As it can be seen in the Figure \ref{fig:dthand}, the center point of the hand has the largest distance (depicted by maximum intensity). Therefore, the pixel having the maximum intensity in the distance transform image (Figure \ref{fig:dthand}), is taken to be the centroid.

The second estimate of the centroid $(x_{c2}, y_{c2})$ is found using the concept of image region moments. The image moments $M_{ij}$, for an image with pixel intensities $I(x, y)$, can be calculated as:
\begin{equation}
M_{ij} = \sum_x^{}\sum_y^{}x^{i}y^{j}I(x ,y)
\end{equation}
The centroid $(x_{c2}, y_{c2})$ is given by:
\begin{equation}
x_{c2} = \frac{M_{10}}{M_{00}} , \; y_{c2} = \frac{M_{01}}{M_{00}}
\end{equation}
where  $M_{10}, M_{01}, M_{00}$ are the moments of the binary hand region obtained from the previous segmentation step.

The final centroid $(x_{c}, y_{c})$ is obtained as the mean of the two initial estimates.

\subsubsection{Counting Raised Fingers}
We use the geometrical properties of the hand to count the number of raised fingers. From the centroid of the hand $(x_{c}, y_{c})$ we draw a circle of maximum radius such that it lies entirely within the hand i.e., not including any background pixel. We call this circle as the inner circle of maximum radius and take its radius to be $r$. Now with the same center, we construct another circle of radius $R  = M_{r} \times r$ to intersect all the raised fingers. Here, $M_{r}$ is a magnification factor which is chosen to be $1.5$ based on the geometrical shape of the hand and extensive experimentation \citep*{wu2016robust}.

We say that the circle of radius $R$ intersects the hand whenever it crosses from a background pixel to a foreground pixel and vice-versa. Let the number of such intersections be $n$. Therefore, taking into account the fact that the circle intersects each finger twice and excluding the two intersections with the wrist, the number of raised fingers is given by:
\begin{equation}
N = \frac{n}{2} - 1
\end{equation}

Finally, if $N = 1$ i.e., there is a single raised finger, the writing hand pose is detected and the air-writing is initialized, otherwise, the screen is cleared and the entire writing hand pose detection algorithm is repeated on successive frames until a match of $N = 1$ is found.

\subsection{Fingertip Detection} \label{fingertip detection}

Although recent CNN-based approaches can give good results for many object detection tasks, detecting the fingertip directly in RGB videos remains challenging, primarily due to its extremely small dimension. Therefore, following a two-stage pipeline, we first detect and segment the hand as discussed before, and secondly we find the fingertip position from the segmented hand using a new signature function called distance-weighted curvature entropy.

The first step involves the detection of the hand using the Faster R-CNN based hand detector followed by the hand region segmentation as already described in the previous section. Similarly, as discussed before, we calculate the coordinates of the centroid of the segmented binary hand region $(x_{c}, y_{c})$.

A signature function is a 1-D functional representation of a boundary and can be generated in many ways. The basic idea behind this is to reduce the original 2-D boundary representation to a 1-D function, that is easier to describe. We propose a novel signature function called as distance-weighted curvature entropy which is a combination of the distance of each of the contour points of the segmented binary hand region from the centroid and the curvature entropy at each contour point. Fingertips are high curvature points and are also distant from the hand center. We use these two distinctive features of fingertips for their accurate detection.

\begin{figure}[t]
 
\begin{subfigure}{0.5\textwidth}
\includegraphics[scale=1, width=1\linewidth]{and_189.png} 
\caption{}
\label{fig:189_handmask}
\end{subfigure} 
\begin{subfigure}{0.5\textwidth}
\includegraphics[scale=1, width=1\linewidth]{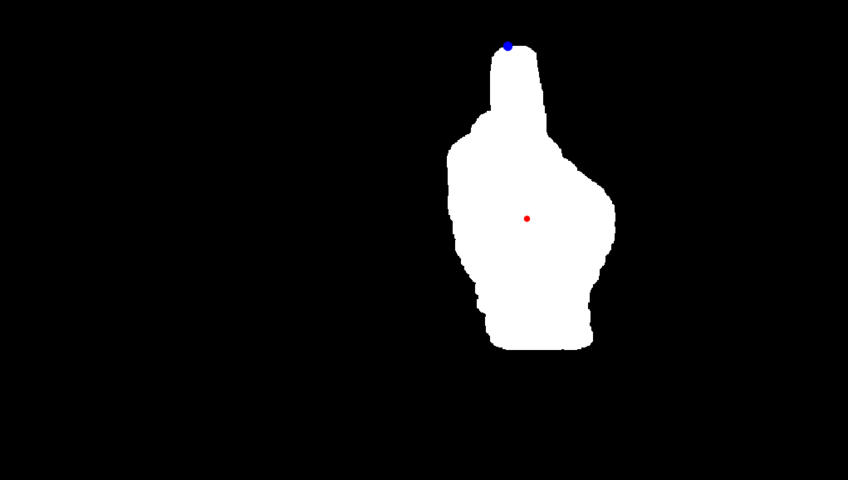}
\caption{}
\label{fig:189_fingertip}
\end{subfigure}

\caption{Fingertip detection. (a) Binary hand image. (b) Binary hand image with centroid and fingertip.}
\label{fig:fingertip_detection}
\end{figure}

The first factor of the proposed signature function is a scale in-variant measure of the curvature of a contour presented in \citep*{feldman2005information} and is referred to as curvature entropy $u$. As shown in Figure \ref{fig:189_handmask}, the segmented hand is represented as a binary image. Let $s$ be a planar contour curve extracted from the segmented hand region of the binary image. If the contour $s$ is of length $L$ and is sampled at $n$ uniformly spaced points, then the sampling interval will be $\triangle s = L/n$. We consider that between any two points along the sampled contour curve, the tangent direction changes by an angle $\alpha$, called the \textit{turning angle}. The curvature $\kappa$ is the change in the tangent direction as we move along the curve and hence may be approximated by the ratio between the turning angle $\alpha$ and the sampling interval $\triangle s$:
\begin{equation}
\kappa(s) \approx \frac{\alpha}{\triangle s}
\end{equation}

Following the derivation in \citep*{feldman2005information}, the curvature entropy $u$ may be approximated as follows:
\begin{equation}
u(\kappa(s)) \propto -\cos(\triangle s \cdot \kappa(s))
\end{equation}
Therefore, we find that the curvature entropy $u$ is is locally proportional to the curvature $\kappa(s)$ and is also scale-invariant. This allows us to localize high curvature points, namely the fingertip in our case, without the need for exploring different scales.

The second factor of the signature function will be a distance function $\delta(s)$ which is equal to the distance between each contour point and the centroid of the hand $(x_{c}, y_{c})$.

We define the signature function of a contour $\Psi(s)$ as the product of the curvature entropy of the contour $u(\kappa(s))$ and the distance function $\delta(s)$:
\begin{equation}
\Psi(s) = u(\kappa(s)) \cdot \delta(s)^{\gamma}
\end{equation}
where the parameter $\gamma$ is a weight for the distance term in the signature function and is tuned during experiments. The distance term weighted by the parameter $\gamma$ eliminates any high curvature points along the hand contour that are not fingertips, thereby reducing false positives caused by irregularities in the hand contour.

Since the fingertip will have a high value of curvature entropy and will be also faraway from the hand centroid, the signature function $\Psi(s)$, which may be considered as an 1-D array, will have a maximum at the fingertip position (see Figure \ref{fig:189_fingertip}). Therefore, the coordinates of the fingertip $(x_{f}, y_{f})$ may be calculated as follows:
\begin{equation}
(x_{f}, y_{f}) = s(\operatorname*{arg \, max}_s \Psi(s))
\end{equation}

\subsection{Fingertip Tracking and Character Trajectory Generation} \label{fingertip tracking}
The third step in the air-writing recognition system is the generation of characters from trajectories formed by sequentially detected fingertip locations. For this, we need to track the fingertips over successive frames starting from the frame in which the writing hand pose is detected until a stopping criterion is reached. This will generate the required character trajectory. Following this, we apply a trajectory smoothing algorithm to remove noise on account of poor fingertip detection or hand trembling.

\subsubsection{Fingertip Tracking} 
Detection and tracking of the hand over successive frames is integral to the fingertip detection and tracking performance. Experiments show that using the Faster R-CNN hand detector for every frame is computationally costly and leads to frame rates much lower than real-time performance. Therefore, the KCF tracking algorithm \citep*{henriques2015high} is used for tracking of the detected hand region. The tracker is initialized with the Faster R-CNN hand detector output and re-initialization is done at an interval of $t$ frames. Re-initialization is necessary in this case since the hand being a highly non-rigid object is difficult to track over a long time. Taking the interval for re-initialization $t$ to be equal to 50 is experimentally found to give the best compromise between tracking accuracy and frame rate.

Tracking a fingertip robustly and accurately is fairly challenging due to its small dimension. However, it is essential to an air-writing application since any erroneous track results in a distorted character. Following the success of CNN-based detection-tracking of fingertips in \citep*{huang2016pointing}, we use a detection-based fingertip tracking approach. In each frame, once the hand region is obtained, fingertip detection is carried out as discussed in Section \ref{fingertip detection}. Experiments prove that the proposed detection-based tracking method gives the best result in terms of tracking accuracy and speed while state-of-the-art tracking algorithms either suffer from poor frame rate or the fingertip track is lost after only a few frames.

\subsubsection{Termination Criterion}
The termination or delimiting criterion is important to an air-writing system since there is no pen-up motion, as in case of traditional online handwriting. Once the termination criterion is satisfied, the character trajectory is assumed to be complete and the trajectory is passed to the following smoothing and recognition stages. We use the velocity of the detected fingertip as the natural stopping criterion for the air-writing recognition system. The rationale behind this choice is that the velocity of the fingertip while writing the character will be high compared to when the writing is complete and the fingertip will be nearly at rest. Let the coordinates of fingertip for the $r^{th}$ frame be $(x_{r}, y_{r})$, and the coordinates for the $(r+1)^{th}$ frame be $(x_{r+1}, y_{r+1})$. Then the displacement $d$ of the fingertip over one frame is given by the Euclidean distance between the two points and the velocity $v$ is the product of the displacement and the frame rate $f$ in fps as shown in the following equations:

\begin{equation}
d = \sqrt{(x^{r+1} - x^{r})^2 + (y^{r+1} - y^{r})^2}
\end{equation}

\begin{equation}
v = d \cdot f
\end{equation}

When the velocity $v < \tau$ the stopping criterion is satisfied. Here $\tau$ is a threshold and experiments reveal the a value of $\tau = 40$ gives best results for most users. Besides, in case a distorted trajectory is plotted on account of hand trembling or any other disturbance, the user can manually reach the stopping criterion by providing a non-writing hand pose, which also clears the screen and then again initialize the writing sequence by a writing hand pose in following frames.

\begin{figure}[t]
 
\begin{subfigure}{0.24\textwidth}
\includegraphics[scale=1, width=1.1\linewidth, height=2.4cm]{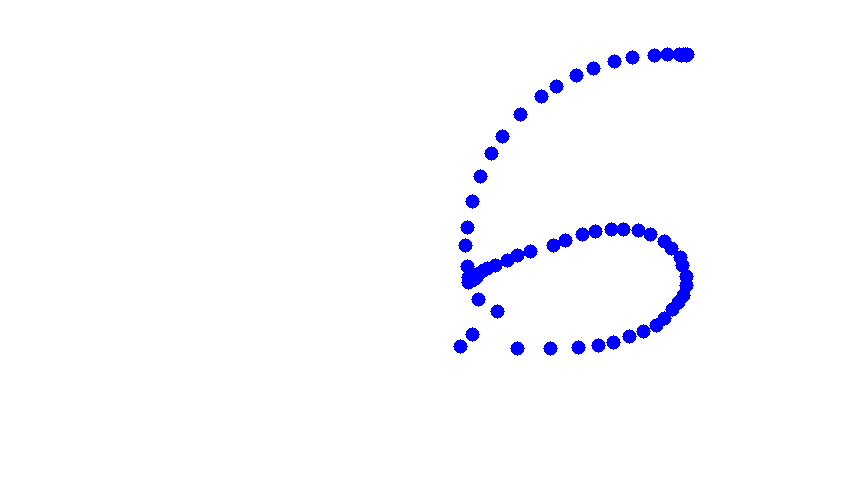} 
\caption{Initial trajectory.}
\label{fig:subim1}
\end{subfigure}
\begin{subfigure}{0.24\textwidth}
\includegraphics[scale=1, width=1.1\linewidth, height=2.4cm]{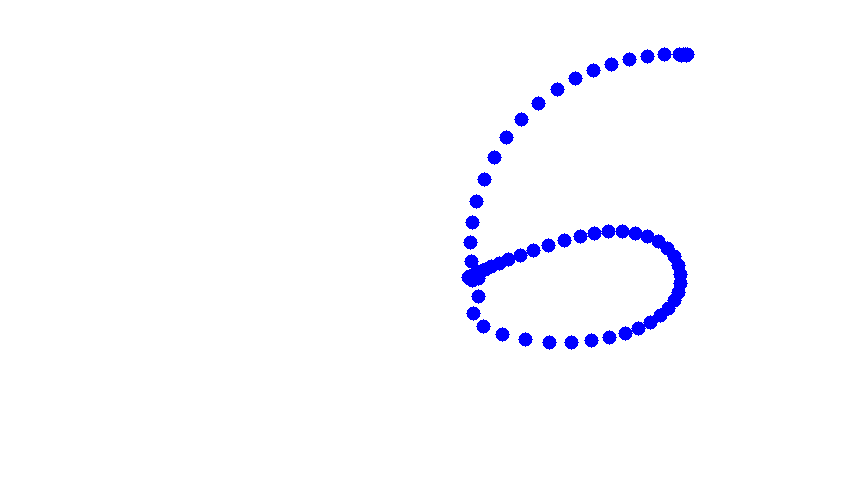}
\caption{First iteration.}
\label{fig:subim2}
\end{subfigure}
\begin{subfigure}{0.24\textwidth}
\includegraphics[scale=1, width=1.1\linewidth, height=2.4cm]{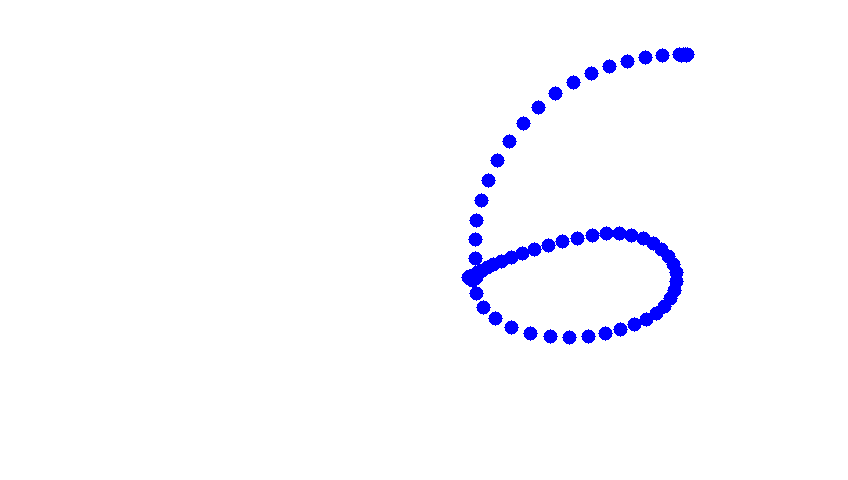} 
\caption{Second iteration.}
\label{fig:subim1}
\end{subfigure}
\begin{subfigure}{0.24\textwidth}
\includegraphics[scale=1, width=1.1\linewidth, height=2.4cm]{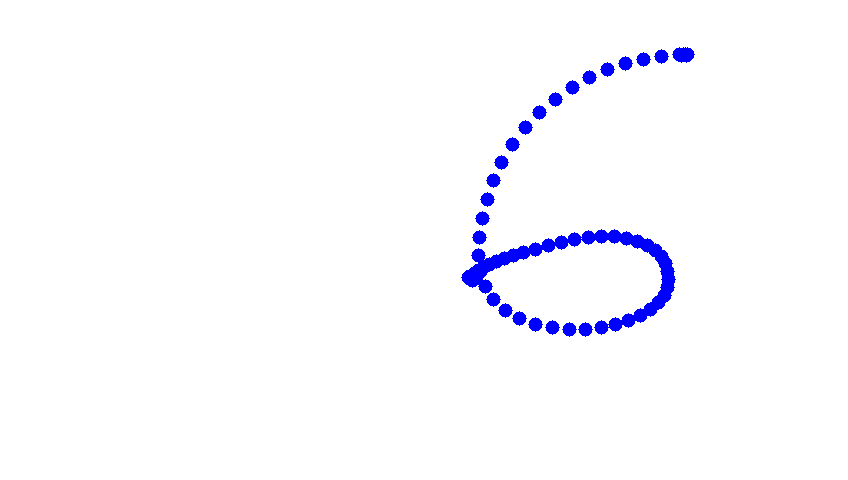}
\caption{Final iteration.}
\label{fig:subim2}
\end{subfigure}
 
\caption{Trajectory smoothing for an initially distorted character trajectory.}
\label{fig:smoothing}
\end{figure}

\subsubsection{Trajectory Smoothing}
Poor fingertip detection or trembling of hands may lead to the formation of uneven or distorted character trajectory, resulting in misclassification of characters in the recognition stage. Thus, smoothing of the character trajectory is an essential post-processing step in order to attenuate the noise.

Let $T$ be the character trajectory obtained from the preceding steps. We propose a simple iterative smoothing algorithm that replaces a trajectory point $T_{i}$ by the average value of the two neighboring points $T_{i-1}$ and $T_{i+1}$, based on the condition that the distance between the point $T_{i}$ and its preceding point $T_{i-1}$ is greater than a threshold $\lambda$. The process is repeated over the entire trajectory until the difference between curvature entropy of the trajectory $u(T)$  for two consecutive iterations is less than a tolerance $\epsilon$. The stopping criterion for the iteration is based on the concept of curvature entropy of a curve described in Section \ref{fingertip detection}. It is found experimentally that the average distance between two consecutive points of a character trajectory lies close to 1. Therefore, the value of $\lambda$ is experimentally chosen as 5, to give a trade-off between accuracy and speed. Also, experiments show that the tolerance $\epsilon$ kept at 0.4 gives fairly a good performance in real-time. The proposed smoothing algorithm gives competitive results compared to others, namely the Ramer-Douglas-Peucker algorithm \citep*{misra2017vision} at a significantly lower computational cost. Visualizations for successive iterations involved in smoothing of an initially distorted character trajectory is shown in Figure \ref{fig:smoothing}.

\subsection{Character Recognition} \label{character recognition}

\begin{figure}[t]
 
\begin{subfigure}{0.5\textwidth}
\includegraphics[scale=1, width=1\linewidth]{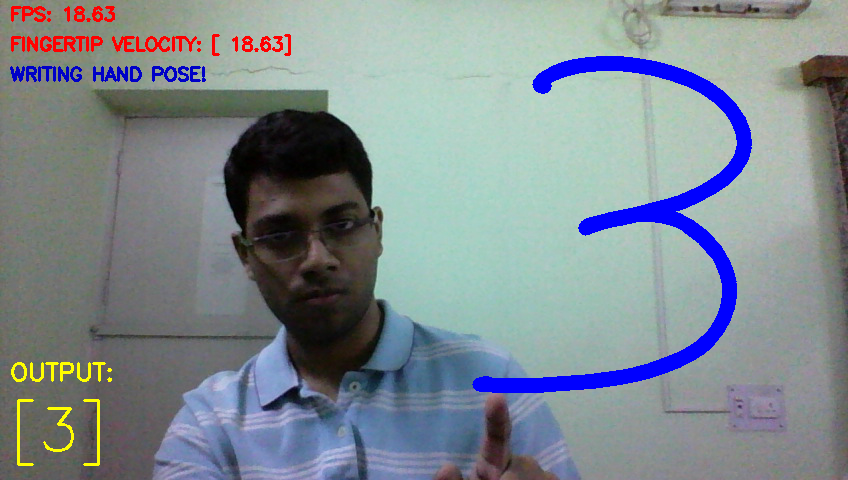} 
\caption{}
\label{fig:subim1}
\end{subfigure} 
\begin{subfigure}{0.5\textwidth}
\includegraphics[scale=1, width=1\linewidth]{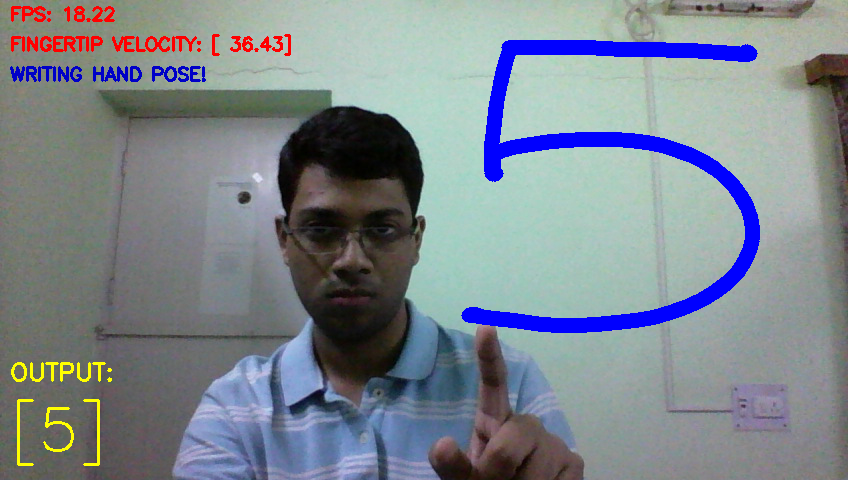}
\caption{}
\label{fig:subim2}
\end{subfigure}
\begin{subfigure}{0.5\textwidth}
\includegraphics[scale=1, width=1\linewidth]{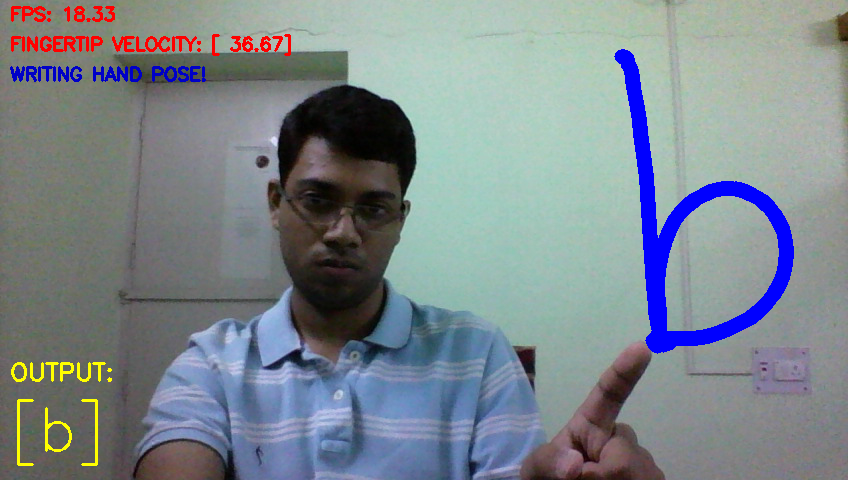} 
\caption{}
\label{fig:subim3} 
\end{subfigure}
\begin{subfigure}{0.5\textwidth}
\includegraphics[scale=1, width=1\linewidth]{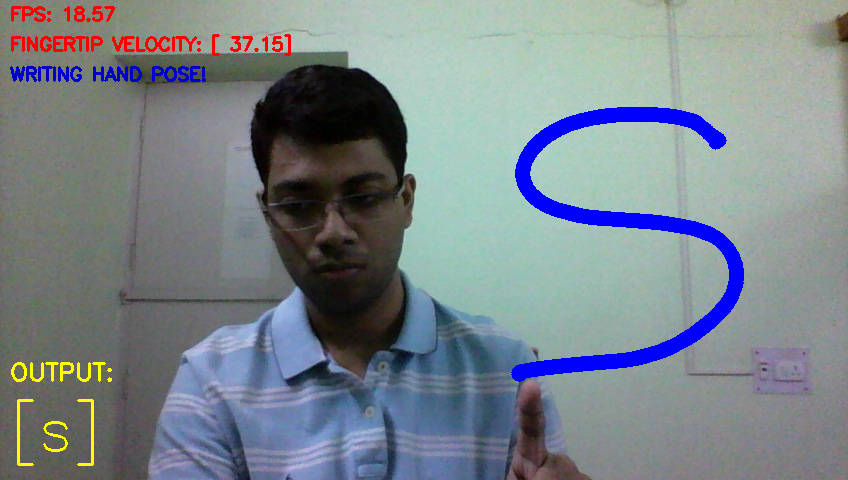} 
\caption{}
\label{fig:subim4}
\end{subfigure} 

\caption{Air-writing character trajectories of digits and letters along with hand pose description, predicted outputs and frame rates using the proposed air-writing recognition system.}
\label{fig:finalopimages}
\end{figure}

Following the recent success of deep convolutional neural networks (CNNs) on handwritten character recognition tasks, we use the AlexNet architecture \citep*{krizhevsky2012imagenet}, pre-trained on the EMNIST dataset \citep*{cohen2017emnist} for the recognition of the air-writing character trajectories. Figure \ref{fig:finalopimages} shows the final character recognition results for four different characters using the proposed framework. 

\section{Experiments} \label{experiments}
\begin{figure}[t]
 
\begin{subfigure}{0.5\textwidth}
\includegraphics[scale=1, width=1\linewidth]{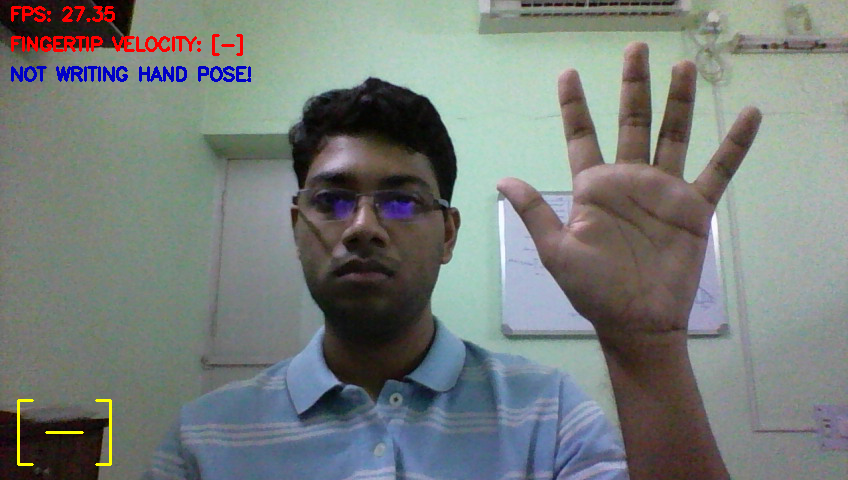} 
\caption{}
\label{fig:demo1}
\end{subfigure} 
\begin{subfigure}{0.5\textwidth}
\includegraphics[scale=1, width=1\linewidth]{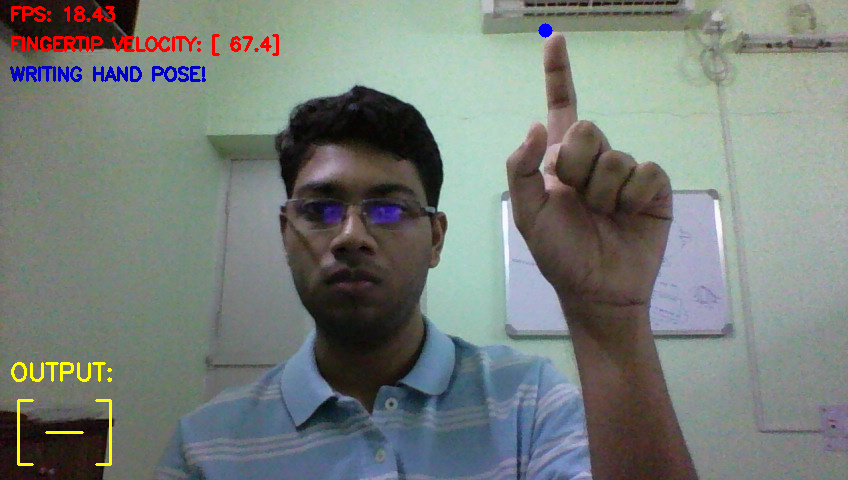}
\caption{}
\label{fig:demo2}
\end{subfigure}
\begin{subfigure}{0.5\textwidth}
\includegraphics[scale=1, width=1\linewidth]{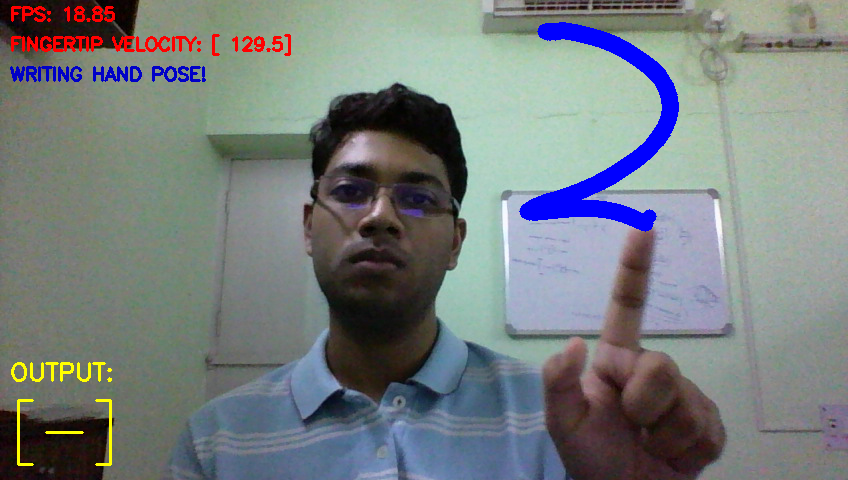} 
\caption{}
\label{fig:demo3} 
\end{subfigure}
\begin{subfigure}{0.5\textwidth}
\includegraphics[scale=1, width=1\linewidth]{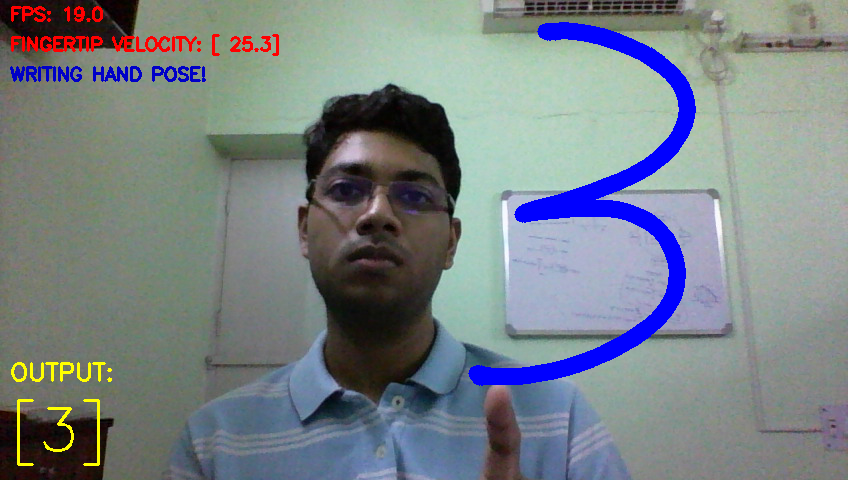} 
\caption{}
\label{fig:demo4}
\end{subfigure} 

\caption{Different stages of air-writing using the proposed framework. (a) Non-writing hand pose, therefore, fingertip detection does not occur. (b) User starts writing, writing hand pose is detected. The fingertip is detected and tracked over successive frames. (c) User is in the process of writing a character. Character recognition does not occur since the termination criterion is not satisfied.  (d) Fingertip velocity is less than the threshold. Therefore, the termination criterion is satisfied, and the character '3' is recognized.}
\label{fig:demoimg}
\end{figure}

In this section, we present extensive experiments to demonstrate the superior performance of the proposed fingertip detection and tracking algorithm and state-of-the-art recognition performance for air-writing character trajectories. All experiments are performed on a machine equipped with a single NVIDIA GeForce GTX 1080 and an Intel Core i7-4790K Processor with 16GB RAM. Figure \ref{fig:demoimg} shows the different stages involved in performing a complete air-writing gesture by an user using the proposed framework. 

\subsection{Dataset Preparation}
The dataset contains air-writing videos with corresponding fingertip trajectories by five subjects. The trajectories have been recorded for numbers (from 0 to 9) and English alphabet letters (from a to z) using a standard web-cam with backgrounds of varying complexity and illumination levels. In total, 1800 different air-writing character trajectories (10 samples each of 36 characters) have been recorded. Each trajectory consists of a set of points $(x, y, t)$, where $(x, y)$ is the position of the fingertip at time $t$ . Our dataset differs from egocentric fingertip detection datasets viz. the SCUT EgoFinger Dataset \citep*{huang2016pointing} in the fact that the video sequences have been captured using a web-cam and hence contains the subject's face as well (which is mostly absent in case of egocentric videos). This makes the task more challenging.

\begin{table}[t]
\centering
\resizebox{\textwidth}{!}{\begin{tabular}{@{}llllll@{}}
\toprule
\multirow{2}{*}{\textbf{Tracker}} & \multicolumn{2}{c}{\textbf{OPE}} & \multicolumn{2}{c}{\textbf{TRE}} & \multirow{2}{*}{\textbf{Speed (fps)}} \\
                         & IoU      & Precision    & IoU      & Precision    &                              \\ \midrule
KCF \citep*{henriques2015high}     & \underline{58.7}     & \underline{71.6}         & \underline{60.4}     & \underline{73.5}         & \textbf{\underline{27.1}}                         \\
TLD \citep*{kalal2012tracking}     & \textbf{60.2}     & \textbf{74.1}         & \textbf{62.5}     & \textbf{76.3}         & 14.5                         \\
MIL \citep*{babenko2011robust}     & 48.7     & 58.4         & 49.7     & 58.9         & 16.1                         \\ \bottomrule
\end{tabular}}
\caption{Hand tracking performance on the test set in terms of overlap (IoU) and precision following the OTB-2013 benchmark. The \underline{proposed} and \textbf{best} results are highlighted for each column.}
\label{table:hand}
\end{table}

\subsection{Hand Detection and Tracking} \label{hand experiment}

It is essential to evaluate the hand detection and tracking performance, since it forms the first stage of the algorithm. As explained in Section \ref{hand detection}, the Faster R-CNN framework is used for the detection of hands. The proposed Faster R-CNN based hand detector is trained on a dataset consisting of 15,000 images collected from the EgoFinger dataset \citep*{huang2016pointing} and the EgoHands dataset \citep*{bambach2015lending} with annotated hand regions. We use Inception-v2 model \citep*{szegedy2016rethinking} trained on the Microsoft COCO dataset \citep*{lin2014microsoft} as the base network for Faster R-CNN. The network is trained end-to-end using stochastic gradient descent with momentum. The momentum value used  is 0.9 and the learning rate starts from 0.0002 and decreases to 0.00002 after 900,000 steps.

For hand tracking, we compare three state-of-the-art tracking algorithms, viz.  KCF \citep*{henriques2015high}, TLD \citep*{kalal2012tracking} and MIL \citep*{babenko2011robust}, initialization in each case being done with the detected hand regions using FRCNN. As pointed out earlier re-initialization of the tracker is done at an interval of 50 frames to enable long-term tracking. We test the hand tracking performance for 1200 video sequences from our air-writing dataset. 

We use the OTB-2013 \citep*{wu2013online} benchmark to evaluate our hand tracking results. The OTB-2013 benchmark considers the average per-frame success rate at different values of thresholds. For a particular frame, if the intersection-over-union (IoU) between the estimate produced by a tracker and the ground-truth is greater than a certain threshold, the tracker is successful in that frame. For comparing different trackers, we use the area under the curve of success rates for different values of the threshold. The overlap (IoU) and precision scores for OPE (one pass evaluation) and TRE (temporal robustness evaluation) have been reported  in Table \ref{table:hand}. In OPE, the tracker is run once on each video sequence, from the start to the end, while in TRE, the tracker is started from twenty different starting points, and run until the end from each. 

\begin{figure}[t]
 
\begin{subfigure}{0.5\textwidth}
\includegraphics[scale=1, width=1\linewidth]{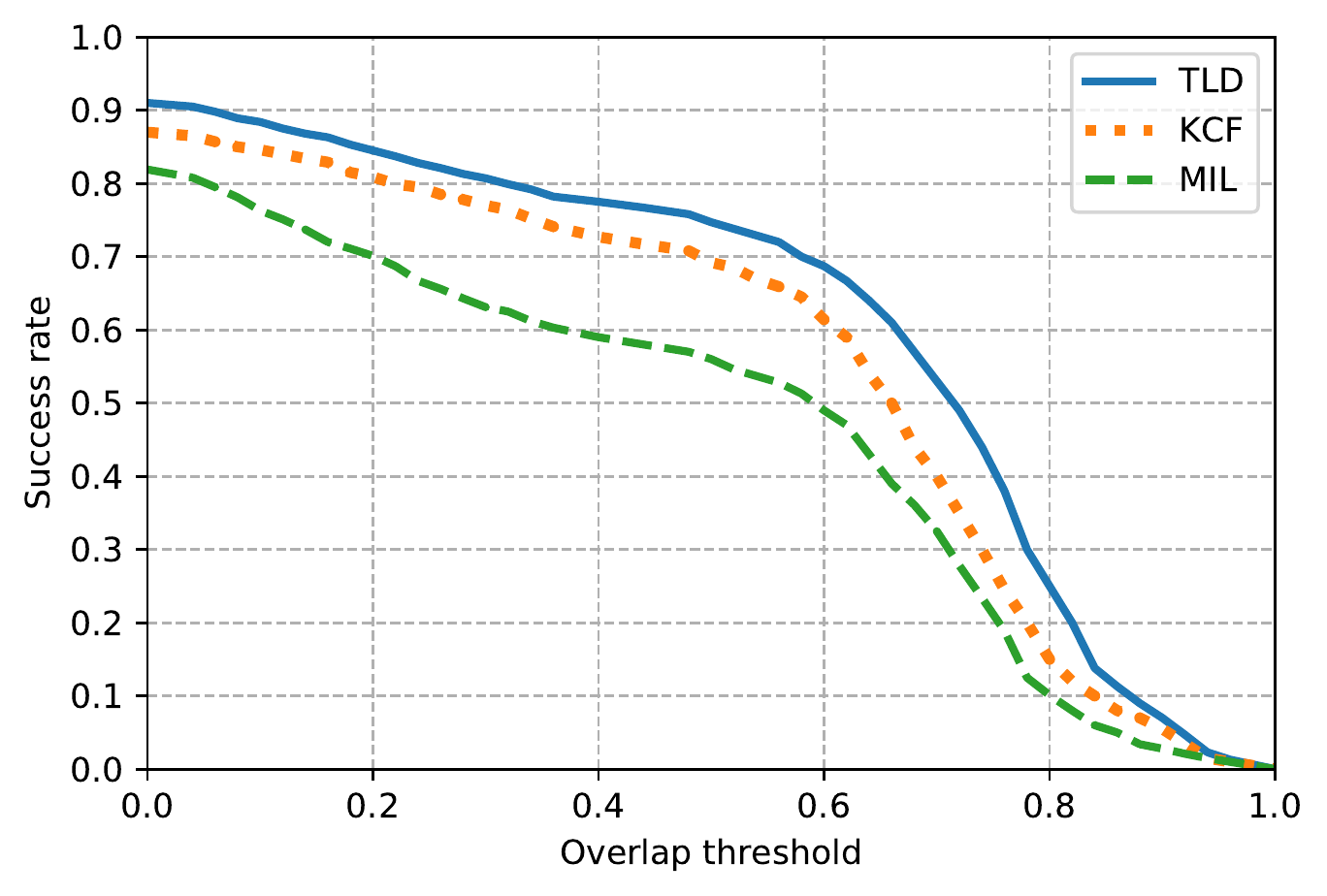} 
\caption{Success plots for OPE.}
\label{fig:s_plot}
\end{subfigure}
\begin{subfigure}{0.5\textwidth}
\includegraphics[scale=1, width=1\linewidth]{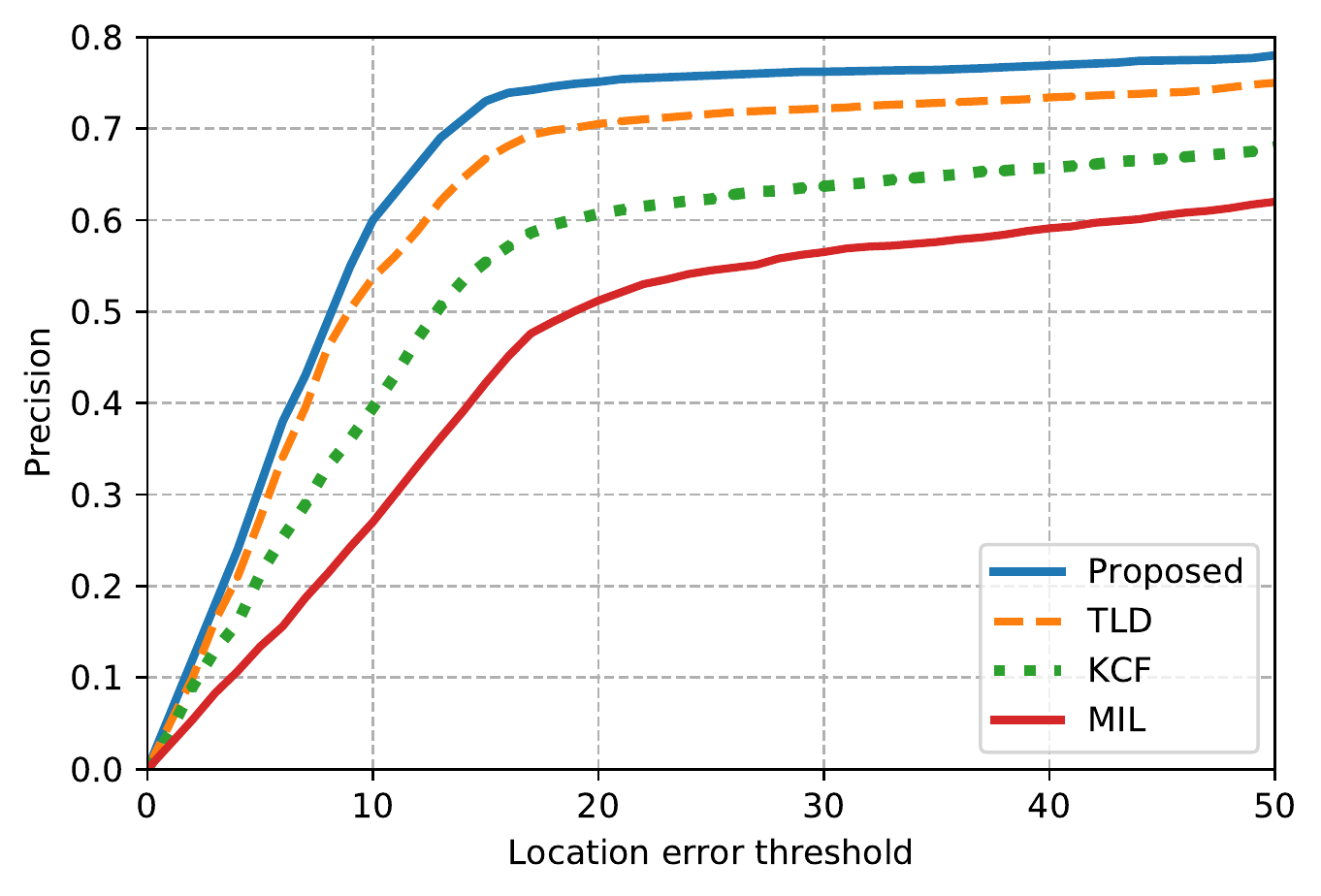}
\caption{Precision plots for OPE.}
\label{fig:p_plot}
\end{subfigure}
 
\caption{Comparison of tracking performances. (a) Hand 	tracking. (b) Fingertip tracking.}
\label{fig:graphimgs}
\end{figure}

While MIL clearly suffers from the drifting problem, TLD gives superior tracking performance in terms of precision but has a very poor frame rate. Experiments on the test set reveal that Faster R-CNN based hand detection followed by KCF tracking gives the best compromise between precision of tracking and frame rate and therefore we use it for fingertip detection. Figure \ref{fig:s_plot} shows the summarized hand tracking performance.

\begin{table}[t]
\centering
\resizebox{\textwidth}{!}{\begin{tabular}{@{}lll@{}}
\toprule
\textbf{Tracker}                 & \multicolumn{1}{c}{\textbf{\begin{tabular}[c]{@{}c@{}}Mean Precision \\ (15 px.)\end{tabular}}} & \textbf{Speed (fps)} \\ \midrule
Proposed (Tracking-by-detection) & \textbf{\underline{73.1}}                                                                                   & \underline{18.5}                 \\
KCF \citep*{henriques2015high}                             & 55.4                                                                                            & \textbf{26.4}        \\
TLD \citep*{kalal2012tracking}                             & 66.7                                                                                            & 10.6                 \\
MIL \citep*{babenko2011robust}                             & 42.4                                                                                            & 12.1                 \\ \bottomrule
\end{tabular}}
\caption{Fingertip tracking performance on the test set in terms of mean precision (15 px.). The \underline{proposed} and \textbf{best} results are highlighted for each column.}
\label{table:fingertip}
\end{table}

\subsection{Fingertip Detection and Tracking}

A test set of 1200 video sequences from our air-writing dataset has been used for  evaluation of fingertip detection and tracking performance. For fingertip detection, the distance weighting parameter $\gamma$, referred to in Section \ref{fingertip detection}, is tuned for maximum detection accuracy. The value of $\gamma$ is varied in the range [1, 5] with steps of 0.5, the maximum detection accuracy occurring at 2.5.    

For fingertip tracking, we use the precision curve proposed in \citep*{henriques2015high} as the evaluation metric. A frame may be considered correctly tracked if the predicted target center is within a distance threshold of the ground truth. Precision curves show the percentage of correctly tracked frames for a range of distance thresholds. This is particularly suitable for our fingertip tracking application since the fingertip is best represented by a point and not a bounding box.

A tracker having higher precision at low thresholds is more accurate, while a lower precision at high thresholds implies that the target is lost. When a representative precision score is needed, we choose 15 pixels as the threshold instead of the standard practice of choosing 20 pixels \citep*{henriques2015high}, since the fingertip tracking performance largely determines the resulting character trajectory, and hence the lower tolerance. The precision scores have been reported in Table \ref{table:fingertip}. The results clearly show the superior performance of the proposed fingertip tracking algorithm over state-of-the-art trackers. Figure \ref{fig:p_plot} shows the summarized fingertip tracking performance.

\subsection{Air-Writing Character Recognition}

\begin{figure*}[t]
\centering
\includegraphics[scale=0.75]{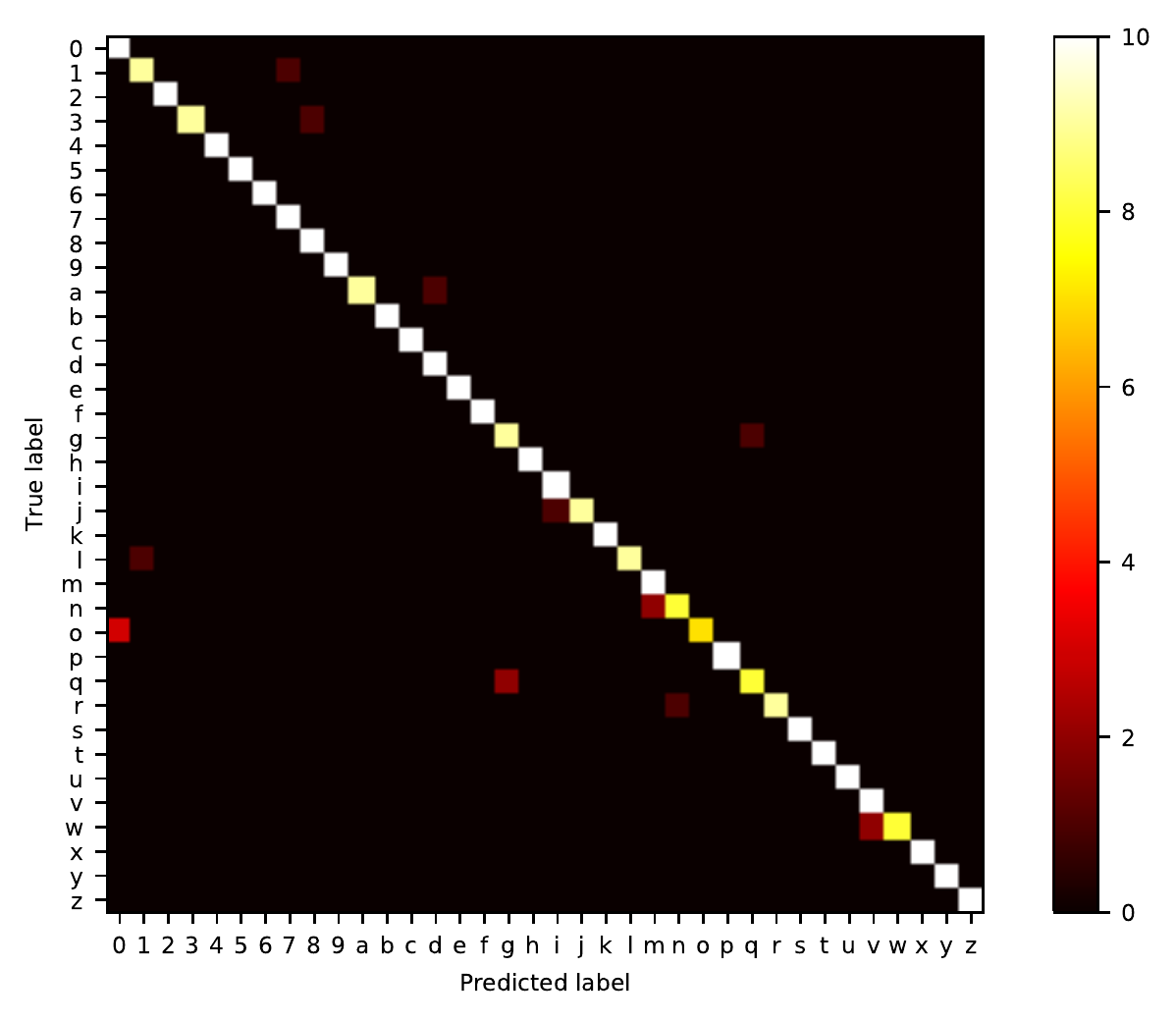}
\caption{Confusion matrix for air-writing character recognition results using the proposed framework.}
\label{fig:cm}
\end{figure*}

For character recognition experiments, the entire air-writing dataset consisting of 1800 video sequences has been used as the test set. The character recognition model based on the AlexNet architecture and pre-trained on the EMNIST dataset gives an air-writing character recognition accuracy of 96.11 \% on the test set. The confusion matrix for character recognition results in Figure \ref{fig:cm} shows that most characters have been correctly recognized, but some characters having similar shape and appearance were confused by the system, such as  \textit{'1'-'7', '3'-'8', 'a'-'d', 'g'-'q', 'j'-'i', 'l'-'1', 'n'-'m', 'o'-'0', 'r'-'n', 'w'-'v'}. Actually is has been found that these similar looking characters are often confused by humans too. The erroneous character recognition results might be improved by better models for handwritten character recognition using context such as words to decide the correct character.

\section{Conclusion} \label{conclusion}

\begin{figure}[t]
 
\begin{subfigure}{0.5\textwidth}
\includegraphics[scale=1, width=1\linewidth]{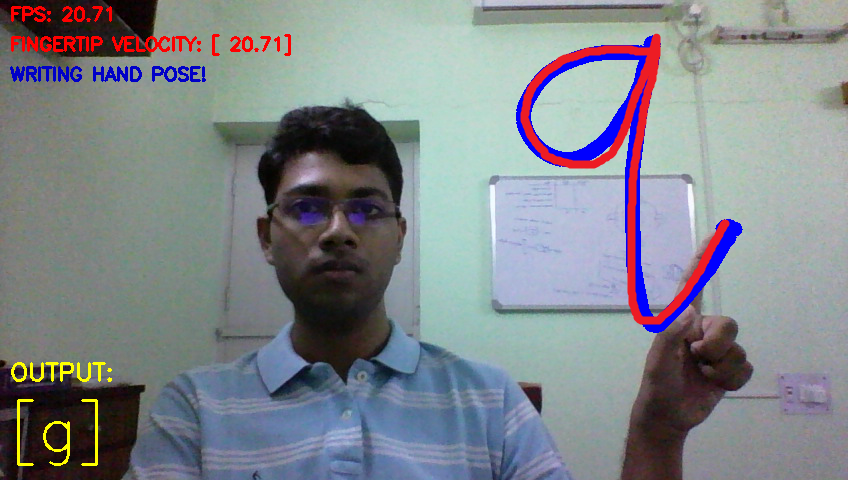} 
\caption{}
\label{fig:failureq}
\end{subfigure} 
\begin{subfigure}{0.5\textwidth}
\includegraphics[scale=1, width=1\linewidth]{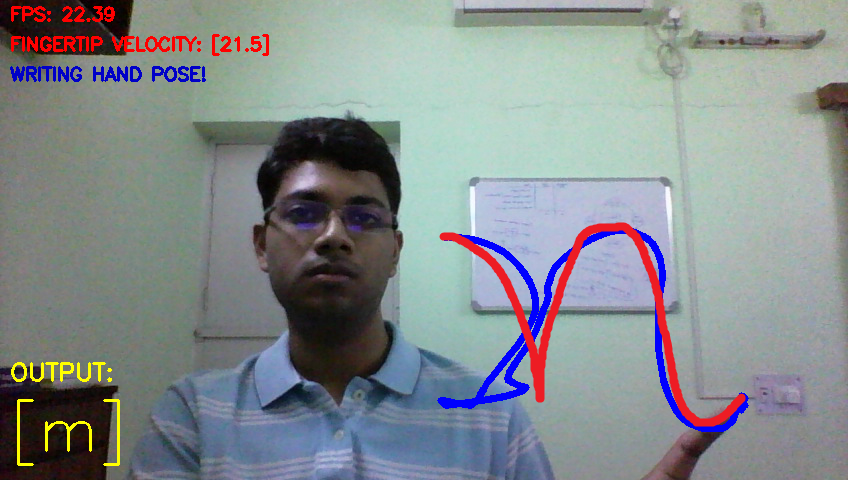}
\caption{}
\label{fig:failuren}
\end{subfigure}

\caption{Failure cases for the proposed air-writing recognition system. (a) Failure in character recognition. (b) Failure in fingertip detection and tracking. The tracked fingertip trajectory is shown in blue color and the ground truth trajectory is shown in red color.}

\label{fig:failurecases}
\end{figure}

In this paper, we presented a new framework for the recognition of mid-air finger writing using web-cam video as input. We proposed a new writing hand pose detection algorithm for the initialization of air-writing. Further, we used a novel signature function called distance-weighted curvature entropy for robust fingertip detection and tracking. Finally, a fingertip velocity based termination criterion was used as a delimiter to mark the completion of the air-writing gesture. Extensive experiments on our air-writing dataset revealed the superior performance of the proposed fingertip detection and tracking approach over state-of-the-art trackers while achieving real-time performance in terms of frame rate. Character recognition results are impressive as well.  

The proposed air-writing recognition framework can find applications in HCI as a virtual touch-less text entry interface. A key application may be in smart home automation for gesture-controlled smart home devices. This is analogous to home automation hubs such as the Amazon Echo, which uses voice commands to control smart home devices. Instead, our framework (with suitable hardware implementation) can be used to take a fingertip trajectory based visual command (using a smart camera) as input to perform a particular task, based on the visual command or keyword. For e.g., a keyword \textit{'b5'} may be used to turn on the fifth light of the bedroom.

Some failure cases however exist for the proposed air-writing recognition system, as depicted in Figure \ref{fig:failurecases}. In Figure \ref{fig:failureq}, it is clearly seen that the failure results due to misclassification by the character recognition model. The failure case in Figure \ref{fig:failuren} is however more subtle, as it results from poor fingertip detection and tracking. This can be accounted for by the fact that, when the hand is not completely inside the frame, the hand detector performance deteriorates. This in turn degrades the fingertip detection and tracking performance. Moreover, when the hand is largely occluded, the geometrical features of the hand used for fingertip detection are also no longer  valid.   

As a future work, we would like to extend the proposed framework to utilize the spatio-temporal as well as motion features of the air-writing trajectories for the recognition of characters. Further, we would also like to extend our framework to the recognition of words as well as signatures, which can have potential application in touch-less and marker-less biometric authentication systems. More robust fingertip detection techniques can also be explored to improve the overall performance of the air-writing recognition system.

\section*{Acknowledgements} 

\noindent\textbf{Funding:} This research did not receive any grant from funding agencies in the public, commercial, or not-for-profit sectors. 
\\
\textbf{Conflict of interest:} The authors declare that there is no conflict of interest regarding the publication of this paper.
\\
\textbf{Informed consent:} Informed consent was obtained from all
individual participants included in the study.


\bibliographystyle{apalike}

\bibliography{egbib}

\end{document}